\documentclass{article}

\usepackage{arxiv}

\usepackage[utf8]{inputenc} 
\usepackage[T1]{fontenc}    
\usepackage{hyperref}       
\usepackage{url}            
\usepackage{booktabs}       
\usepackage{amsfonts}       
\usepackage{nicefrac}       
\usepackage{microtype}      
\usepackage{lipsum}
\usepackage{graphicx}
\graphicspath{ {./images/} }
\usepackage{graphicx,verbatim}
\usepackage{amsmath}
\usepackage{amssymb}
\usepackage{algorithm}
\usepackage{algpseudocode}
\usepackage{booktabs} 
\usepackage{multirow}

\title{SpectralMamba-UNet: Frequency-Disentangled State Space Modeling for Texture-Structure Consistent Medical Image Segmentation}

\author{
 Fuhao Zhang \\
  College of Computer Science\\
  Sichuan Normal University\\
  Chengdu, Sichuan 610068, China \\
  \texttt{fuahozhang@stu.sicnu.edu.cn} \\
   \And
 Lei Liu \\
  Zhejiang University \& Ant Group \\
  Hangzhou, Zhejiang, China \\
  \texttt{liulei1497@gmail.com} \\
  \And
 Jialin Zhang \\
  State Key Laboratory of Public Big Data\\
  Guizhou University\\
  Guiyang, Guizhou 550025, China \\
  \texttt{gs.jlzhang25@gzu.edu.cn} \\
  \And
 Ya-Nan Zhang \\
  College of Computer Science\\
  Sichuan Normal University\\
  Chengdu, Sichuan 610068, China \\
  \texttt{zyn962464@gmail.com} \\
  \And
 Nan Mu $^{*}$ \\
  College of Computer Science\\
  Sichuan Normal University\\
  Chengdu, Sichuan 610068, China \\
  \texttt{nanmu@sicnu.edu.cn} \\
}

\begin{document}
\maketitle
\begin{abstract}
Accurate medical image segmentation requires effective modeling of both global anatomical structures and fine-grained boundary details. Recent state space models (\textit{e.g.}, Vision Mamba) offer efficient long-range dependency modeling. However, their one-dimensional serialization weakens local spatial continuity and high-frequency representation. To this end, we propose SpectralMamba-UNet, a novel frequency-disentangled framework to decouple the learning of structural and textural information in the spectral domain. Our Spectral Decomposition and Modeling (SDM) module applies discrete cosine transform to decompose low- and high-frequency features, where low frequency contributes to global contextual modeling via a frequency-domain Mamba and high frequency preserves boundary-sensitive details. To balance spectral contributions, we introduce a Spectral Channel Reweighting (SCR) mechanism to form channel-wise frequency-aware attention, and a Spectral-Guided Fusion (SGF) module to achieve adaptively multi-scale fusion in the decoder. Experiments on five public benchmarks demonstrate consistent improvements across diverse modalities and segmentation targets, validating the effectiveness and generalizability of our approach.
\end{abstract}


\section{Introduction}

Accurate medical image segmentation is crucial for quantitative diagnosis, treatment planning, and disease monitoring. Convolutional neural networks (CNNs), such as U-Net~\cite{ronneberger2015u}, have long served as the backbone of medical segmentation due to their strong locality bias and hierarchical feature learning. However, their limited receptive fields hinder global context modeling, leading to structural inconsistencies, especially in cases with large anatomical variations or pathological deformations.

To overcome this limitation, Vision Transformers (ViTs)~\cite{dosovitskiy2020image} and state space models (SSMs)~\cite{gu2021efficiently,gu2024mamba} have been increasingly adopted, leveraging long-range dependency modeling through self-attention or selective scanning mechanisms. While effective in capturing global patterns, these approaches typically rely on patch tokenization or 1D sequence flattening, which disrupts local spatial coherence and introduces artifacts in high-frequency regions such as organ boundaries and tissue edges. More critically, they treat all spatial frequencies uniformly, failing to distinguish between low-frequency structural priors and high-frequency textural details. This entanglement leads to a trade-off: aggressive global modeling may smooth out critical boundary cues, while preserving local details often comes at the cost of contextual consistency. Recent studies show that high-frequency components are particularly vulnerable during long-sequence processing in SSMs~\cite{zhu2024vision}, further exacerbating this issue.

To this end, we propose SpectralMamba-UNet, a novel frequency-disentangled state space framework that explicitly separates structural and textural information in the spectral domain. By applying the Discrete Cosine Transform to intermediate features, our method decomposes them into low- and high-frequency components, which are processed separately to preserve complementary characteristics: low frequencies model global anatomical layouts via band-wise state space modeling, while high frequencies retain critical edge and texture details. The proposed Spectral Decomposition and Modeling (SDM) module enables efficient long-range reasoning within each frequency band. Further, a Spectral Channel Reweighting (SCR) module adaptively recalibrates channel-wise importance based on frequency-specific statistics, and a Spectral-Guided Fusion (SGF) module injects these spectral priors into the decoder to facilitate frequency-consistent multi-scale integration.


Our key contributions are threefold:

(1) We propose SpectralMamba-UNet, the first framework to integrate frequency disentanglement with state space modeling in medical image segmentation, enabling separate and effective modeling of global structures (via low-frequency components) and fine boundaries (via high-frequency details).

(2) We introduce three key modules (SDM, SCR, and SGF) for spectral decomposition, frequency-aware channel reweighting, and decoder-level spectral guidance, forming a coherent pipeline for disentangled representation learning.

(3) We demonstrate consistent performance gains across five diverse medical datasets (Synapse, ACDC, DRIVE, EAT, IA), validating the effectiveness and generalizability of frequency-domain reasoning in segmentation.

\section{Methodology}

\subsection{Motivation and Overview}

\begin{figure*}[ht]
    \centering
    \includegraphics[width=1\textwidth]{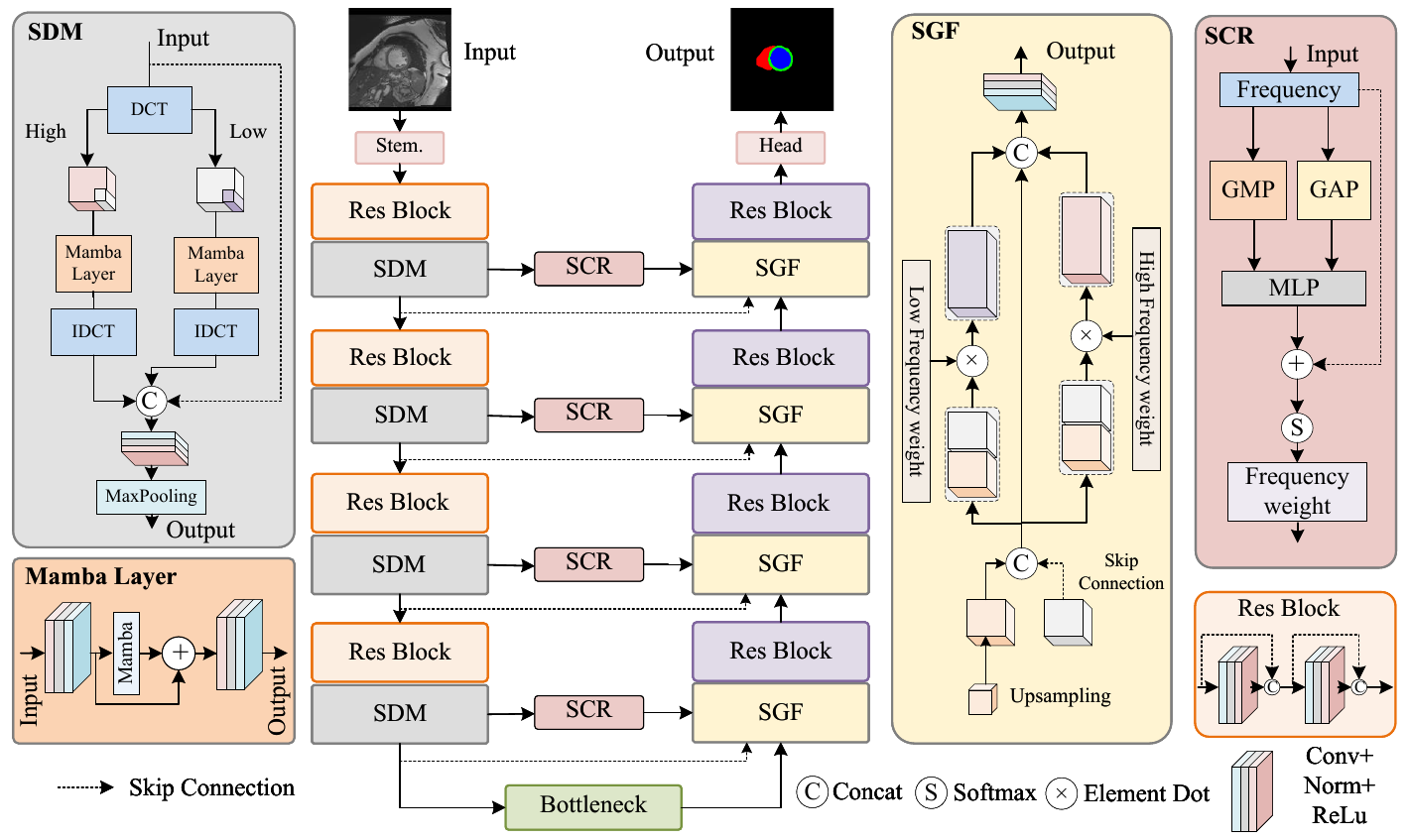}
    \caption{Architecture of SpectralMamba-UNet. SDM performs spectral decomposition, SCR reweights frequency responses, and SGF enables frequency-guided decoder fusion.}
    \label{fig:main_arch}    
\end{figure*}


Conventional segmentation networks primarily operate in the spatial domain, where global anatomical contexts and high-frequency boundary details are implicitly entangled, coupling hampers effective modeling of long-range dependencies and precise capture of fine-grained structural variations.

To address these limitations, we propose SpectralMamba-UNet, a frequency-aware encoder–decoder architecture that integrates spectral decomposition with linear-complexity state space modeling. Concretely, we introduce a spectral-domain representation by applying the Discrete Cosine Transform (DCT) to decompose intermediate features into complementary low- and high-frequency components, enabling explicit frequency disentanglement.

As illustrated in Fig.~\ref{fig:main_arch}, SpectralMamba-UNet comprises three key modules:
(1) Spectral Decomposition and Modeling (SDM) for band-wise feature analysis within the encoder;
(2) Spectral Channel Reweighting (SCR) to perform adaptive, frequency-aware channel recalibration; and
(3) Spectral-Guided Fusion (SGF) to facilitate frequency-conditioned multi-scale feature integration in the decoder. Our architecture explicitly decouples frequency-specific information across the network hierarchy, enhancing both contextual understanding and edge fidelity, while maintaining favorable computational efficiency, making it well-suited for medical image segmentation tasks.


\subsection{Spectral Decomposition and Modeling}
Given an encoder feature map $ X \in \mathbb{R}^{C \times H \times W} $, we apply a 2D Discrete Cosine Transform to project it into the frequency domain, obtaining spectral coefficients $ F = \text{DCT}(X) $. 
Leveraging the energy compaction property of DCT, low-frequency coefficients capture global anatomical structures, while high-frequency coefficients encode fine-grained variations. 
We partition $F$ into complementary low- and high-frequency components using fixed binary masks with ratio $\alpha = 0.125$, yielding $F_{low}$ and $F_{high}$.

To efficiently model long-range dependencies within each band, the spectral maps are reshaped into sequences $S_k \in \mathbb{R}^{C \times (H \cdot W)}$ for $k \in \{low, high\}$ and processed by two independent Mamba blocks:
\begin{equation}
\tilde{S}_k = \text{Mamba}_k(S_k).
\end{equation}

The outputs are reshaped back to spectral maps $\tilde{F}_k$, transformed to the spatial domain via inverse DCT, and fused via convolution with residual addition:
\begin{equation}
X_{SDM} = \text{Conv}(\text{Concat}(X_{low}, X_{high})) + X.
\end{equation}

A fixed spectral partition is adopted for stability and efficiency. 
By explicitly modeling complementary frequency bands, SDM enables frequency-aware representation learning while preserving linear computational complexity.


\subsection{Spectral Channel Reweighting}

Although low- and high-frequency components are modeled separately in SDM, their relative importance may vary across anatomical structures and scales. 
To adaptively balance these responses, we introduce a frequency-conditioned channel reweighting mechanism.

SCR operates on the enhanced spectral representations $\tilde{F}_k$ ($k \in \{low, high\}$). 
For each component, global average pooling (GAP) and global max pooling (GMP) extract complementary channel descriptors. 
The pooled features are passed through a shared multilayer perceptron (MLP), summed, and activated by a sigmoid function to produce channel weights:
\begin{equation}
\mathbf{W}_k =
\sigma\!\left(
\text{MLP}(\text{GAP}(\tilde{F}_k)) +
\text{MLP}(\text{GMP}(\tilde{F}_k))
\right).
\end{equation}

The resulting weights $\mathbf{W}_{low}$ and $\mathbf{W}_{high}$ encode frequency-specific channel importance. 
As spectral and spatial features share aligned channel indexing, these weights are directly propagated to the decoder for frequency-aware modulation.


\begin{algorithm}[t]
\caption{Forward pass of SpectralMamba-UNet}
\label{alg:spectralmamba}
\begin{algorithmic}[1]
\Require Input batch $X \in \mathbb{R}^{B \times C \times H \times W}$
\Ensure Segmentation map $Y_{\mathrm{pred}}$

\State $x\gets X$; $\mathcal{S}\gets\emptyset$; $\mathcal{W}\gets\emptyset$

\For{$k=1$ to $N$}
    \State $x\gets \mathrm{EncBlock}_k(x)$; $F\gets \mathrm{DCT}(x)$; $(F_{\text{low}},F_{\text{high}})\gets \mathrm{MaskSplit}(F,\alpha)$
    \State $\tilde{F}_{\text{low}}\gets \mathrm{Mamba}_{\text{low}}(\mathrm{Flatten}(F_{\text{low}}))$; $\tilde{F}_{\text{high}}\gets \mathrm{Mamba}_{\text{high}}(\mathrm{Flatten}(F_{\text{high}}))$
    \State $x_{\text{sdm}}\gets \mathrm{Conv}(\mathrm{Concat}(
            \mathrm{IDCT}(\tilde{F}_{\text{low}}),
            \mathrm{IDCT}(\tilde{F}_{\text{high}}))) + x$

    \State $(\mathbf{W}^{(k)}_{\text{low}},\mathbf{W}^{(k)}_{\text{high}})
           \gets \mathrm{SCR}(\tilde{F}_{\text{low}},\tilde{F}_{\text{high}})$
    \State $\mathcal{S}\gets \mathcal{S}\cup\{x_{\text{sdm}}\}$;
           $\mathcal{W}\gets \mathcal{W}\cup\{(\mathbf{W}^{(k)}_{\text{low}},
           \mathbf{W}^{(k)}_{\text{high}})\}$

    \State $x\gets \mathrm{Down}_k(x_{\text{sdm}})$
\EndFor

\For{$k=N$ down to $1$}
    \State $x\gets \mathrm{Up}_k(x)$; $(x_{\text{skip}},\mathcal{S})\gets \mathrm{Pop}(\mathcal{S})$
    \State $((\mathbf{W}^{(k)}_{\text{low}},
            \mathbf{W}^{(k)}_{\text{high}}),\mathcal{W})
            \gets \mathrm{Pop}(\mathcal{W})$

    \State $x\gets \mathrm{DecBlock}_k(
            \mathrm{SGF}(\mathrm{Concat}(x,x_{\text{skip}}),
            \mathbf{W}^{(k)}_{\text{low}},
            \mathbf{W}^{(k)}_{\text{high}}))$
\EndFor

\State \Return $\mathrm{Softmax}(\mathrm{Head}(x))$

\end{algorithmic}
\end{algorithm}

\subsection{Spectral-Guided Fusion}

In U-shaped architectures, skip connections typically concatenate encoder and decoder features without accounting for spectral characteristics, which may introduce redundancy across scales.

To incorporate spectral priors during multi-scale fusion, we use the channel weights learned in SCR to modulate skip features. 
Given the upsampled decoder feature $X_{up}$ and encoder feature $X_{skip}$, we concatenate them as $X_{cat} = \text{Concat}(X_{up}, X_{skip})$ and apply frequency-conditioned gating:
\begin{equation}
X_{low}^{g} = X_{cat} \odot \mathbf{W}_{low}, \quad
X_{high}^{g} = X_{cat} \odot \mathbf{W}_{high}.
\end{equation}

The gated features are fused via convolution:
\begin{equation}
X_{out} = \text{Conv}(\text{Concat}(X_{low}^{g}, X_{high}^{g})).
\end{equation}

This channel-aligned spectral modulation promotes frequency-consistent integration between encoder and decoder representations.

\subsection{Overall Forward Process}
Algorithm~\ref{alg:spectralmamba} presents the end-to-end forward propagation of SpectralMamba-UNet, highlighting band-wise spectral modeling, SCR-based frequency weighting, and SGF-guided multi-scale decoding.

\section{Experiments}
\label{sec:experiments}

\subsection{Datasets and Implementation}

We evaluate SpectralMamba-UNet on five public datasets covering diverse modalities and anatomical targets: 
(1) \textbf{Synapse}~\cite{landman2015miccai} (30 abdominal CT scans for multi-organ segmentation); 
(2) \textbf{ACDC}~\cite{bernard2018deep} (100 cardiac MRI scans for RV, myocardium, and LV); 
(3) \textbf{DRIVE}~\cite{staal2004ridge} (40 retinal fundus images for vessel segmentation); 
(4) \textbf{IA}~\cite{mu2023attention} (23 brain CT scans for intracranial aneurysm segmentation); 
(5) \textbf{EAT}~\cite{siriapisith20213d} (220 non-contrast CT scans for epicardial adipose tissue segmentation).

We report Dice Similarity Coefficient (DSC, $\uparrow$) and 95\% Hausdorff Distance (HD95, $\downarrow$). 
All experiments were implemented in PyTorch on an NVIDIA GeForce RTX 3090 GPU. 
Images were resized to $256 \times 256$ and trained for 400 epochs using Adam with an initial learning rate of $1 \times 10^{-4}$. 
Official training–testing splits were adopted for each dataset, and the model was optimized using a combined Dice and cross-entropy loss. 
Unless otherwise stated, the frequency partition ratio was fixed at $\alpha = 0.125$.

\begin{table}[tb]
\centering
\caption{Quantitative comparison on the Synapse dataset (multi-organ CT segmentation). 
Best results are shown in \textbf{bold} and second-best in \underline{underline}.}
\label{tab:sota_synapse}
\resizebox{\textwidth}{!}{%
\begin{tabular}{l|cc|cccccccc}
\toprule
\textbf{Model} & \textbf{DSC}$\uparrow$ & \textbf{HD95}$\downarrow$ 
& \textbf{Aorta} & \textbf{Gallbladder} & \textbf{Kidney (L)} & \textbf{Kidney (R)} 
& \textbf{Liver} & \textbf{Pancreas} & \textbf{Spleen} & \textbf{Stomach} \\
\midrule
Res-UNet   & 77.27 & 20.86 & \underline{88.58} & 55.60 & 80.54 & 77.96 & 94.18 & \underline{61.68} & 88.05 & 71.59 \\
TransUNet & 77.48 & 31.69 & 87.23 & 63.13 & 81.87 & 77.02 & 94.08 & 55.86 & 85.08 & 75.62 \\
Swin-Transformer & 79.13 & 21.55 & 85.47 & \underline{66.53} & 83.28 & 79.61 & \underline{94.29} & 56.58 & \textbf{90.66} & \underline{76.60} \\
UltraLight VM-UNet & 38.12 & 56.28 & 58.96 & -- & -- & -- & 87.69 & 31.50 & 73.48 & 53.35 \\
VM-UNet   & \underline{81.08} & \underline{19.21} & 86.40 & \textbf{69.41} & \textbf{86.16} & \textbf{82.76} & 94.17 & 58.80 & \underline{89.51} & \textbf{81.40} \\
\textbf{SpectralMamba-UNet} & \textbf{81.10} & \textbf{15.31} & \textbf{89.79} & 63.14 & \underline{85.84} & \underline{81.36} & \textbf{94.91} & \textbf{69.69} & 88.86 & 75.09 \\
\bottomrule
\end{tabular}
}
\end{table}

\begin{table}[tb]
\centering
\caption{Quantitative comparison on ACDC (cardiac MRI), DRIVE (retinal vessels), EAT (epicardial adipose tissue), and IA (intracranial aneurysm) datasets. 
Best results are shown in \textbf{bold} and second-best in \underline{underline}.}
\label{tab:comparison_results}
\setlength{\tabcolsep}{3pt}
\resizebox{\columnwidth}{!}{%
\begin{tabular}{lcccccccccc}
\toprule
\multirow{2}{*}{\textbf{Model}} 
& \multicolumn{4}{c}{\textbf{ACDC}} 
& \multicolumn{2}{c}{\textbf{DRIVE}} 
& \multicolumn{2}{c}{\textbf{EAT}} 
& \multicolumn{2}{c}{\textbf{IA}} \\
\cmidrule(lr){2-5} \cmidrule(lr){6-7} \cmidrule(lr){8-9} \cmidrule(lr){10-11}
& \textbf{mDSC}$\uparrow$ & \textbf{RV} & \textbf{Myo} & \textbf{LV} 
& \textbf{DSC}$\uparrow$ & \textbf{HD95}$\downarrow$ 
& \textbf{DSC}$\uparrow$ & \textbf{HD95}$\downarrow$ 
& \textbf{DSC}$\uparrow$ & \textbf{HD95}$\downarrow$ \\
\midrule
Res-UNet  & \underline{91.25} & \underline{90.57} & \underline{89.10} & 94.07 
         & \underline{81.96} & \underline{3.05} 
         & \underline{88.33} & \textbf{1.33} 
         & 84.14 & 34.28 \\
TransUNet & 89.71 & 88.86 & 84.53 & \underline{95.73} 
          & 81.81 & 6.09 
          & 86.36 & 1.64 
          & 80.84 & 61.19 \\
Swin-Transformer & 90.00 & 88.55 & 85.62 & \textbf{95.83} 
          & 71.54 & 4.94 
          & 84.41 & 1.58 
          & 84.49 & \textbf{21.52} \\
UltraLight VM-UNet & 78.84 & 76.68 & 75.53 & 84.32 
                 & 78.57 & 9.64 
                 & 76.78 & 2.76 
                 & \underline{85.23} & 23.90 \\
VM-UNet & 90.56 & 87.77 & 87.89 & 95.02 
        & 80.96 & 8.99 
        & 87.06 & \underline{1.46} 
        & 81.86 & 33.88 \\
\textbf{SpectralMamba-UNet} 
        & \textbf{92.89} & \textbf{92.10} & \textbf{91.39} & 95.18 
        & \textbf{83.61} & \textbf{2.26} 
        & \textbf{90.91} & 2.26 
        & \textbf{86.46} & \underline{23.75} \\
\bottomrule
\end{tabular}
}
\end{table}

\subsection{Comparison with State-of-the-Art}

We compare SpectralMamba-UNet with representative CNN-based (Res-UNet~\cite{xiao2018weighted}), Transformer-based (TransUNet~\cite{chen2021transunet}, Swin-Transformer~\cite{liu2021swin}), and Mamba-based methods (VM-UNet~\cite{ruan2024vm}, UltraLight VM-UNet~\cite{wu2025ultralight}).

\textbf{Quantitative Analysis.}
As shown in Table~\ref{tab:sota_synapse} and Table~\ref{tab:comparison_results}, SpectralMamba-UNet demonstrates consistent improvements across multiple datasets and metrics.
\textbf{(1) Volumetric Structures (Synapse \& ACDC).} 
On Synapse, our method achieves the lowest HD95 (15.31) and competitive mean DSC (81.10\%), with notable gains on challenging organs such as the \textit{Pancreas} (+10.89\% DSC over VM-UNet). 
On ACDC, SpectralMamba-UNet attains the highest mean DSC (92.89\%) and strong performance across all cardiac substructures, including the thin \textit{Myocardium} (91.39\%). 
These results indicate improved structural consistency and boundary delineation.
\textbf{(2) Tubular and Irregular Structures (DRIVE, IA, EAT).} 
For vessel and lesion segmentation, structural continuity is critical. 
SpectralMamba-UNet achieves the best DSC (83.61\%) and lowest HD95 (2.26) on DRIVE, and shows consistent improvements on IA and EAT compared with strong baselines. 
The overall reduction in HD95 suggests more accurate boundary localization for thin or irregular structures.

\begin{figure*}[t]
    \centering
    \includegraphics[width=0.9\textwidth]{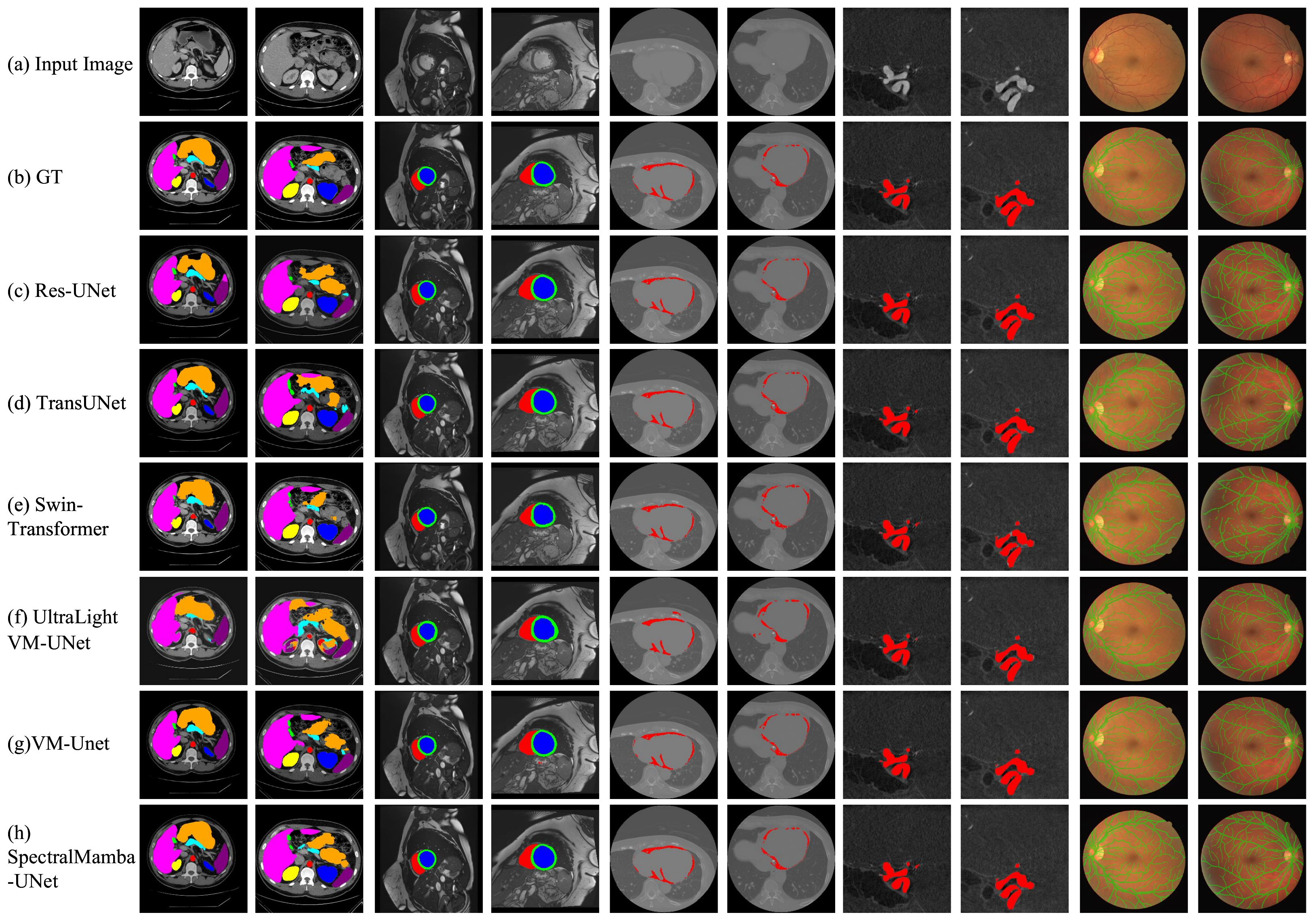} 
    \caption{Qualitative comparison on Synapse, ACDC, EAT, IA, and DRIVE (left to right). Compared with representative baselines (c–g), SpectralMamba-UNet (h) produces sharper boundaries and improved topological consistency.}
    \label{fig:Com_Visual}
\end{figure*}

\textbf{Qualitative Analysis.}
Fig.~\ref{fig:Com_Visual} presents visual comparisons. 
In low-contrast abdominal regions (Synapse), the proposed method produces clearer organ boundaries than Transformer-based baselines. 
For topologically complex structures such as retinal vessels (DRIVE), SpectralMamba-UNet better preserves peripheral connectivity compared with VM-UNet. 
These observations are consistent with the quantitative improvements.

\subsection{Ablation Studies}

\begin{table}[tb]
\centering
\caption{Ablation study of different components. 
+Freq denotes spectral-domain convolution (without Mamba), +Spatial Mamba denotes spatial-domain Mamba modeling, and +Freq+SCR+SGF represents spectral decomposition with channel reweighting and guided fusion. Best results are shown in \textbf{bold}.
}
\label{tab:ablation}
\setlength{\tabcolsep}{3pt}
\resizebox{\textwidth}{!}{%
\begin{tabular}{lcccccccccc}
\toprule
\multirow{2}{*}{\textbf{Variant}} 
& \multicolumn{2}{c}{\textbf{Synapse}} 
& \multicolumn{2}{c}{\textbf{ACDC}} 
& \multicolumn{2}{c}{\textbf{DRIVE}} 
& \multicolumn{2}{c}{\textbf{EAT}} 
& \multicolumn{2}{c}{\textbf{IA}} \\
\cmidrule(lr){2-3} \cmidrule(lr){4-5} \cmidrule(lr){6-7} \cmidrule(lr){8-9} \cmidrule(lr){10-11}
& DSC$\uparrow$ & HD95$\downarrow$ 
& DSC$\uparrow$ & HD95$\downarrow$ 
& DSC$\uparrow$ & HD95$\downarrow$ 
& DSC$\uparrow$ & HD95$\downarrow$ 
& DSC$\uparrow$ & HD95$\downarrow$ \\
\midrule
Baseline & 77.27 & 20.86 & 91.25 & -- & 81.96 & 3.05 & 88.33 & 1.33 & 84.14 & 34.28 \\
+Freq & 77.76 & 17.08 & 91.83 & -- & 83.05 & 2.70 & 88.87 & 1.71 & 85.63 & \textbf{22.76} \\
+Spatial Mamba & 77.51 & 22.07 & 91.76 & -- & 83.07 & 2.58 & 88.40 & 1.54 & 84.66 & 33.03 \\
+Freq+SCR+SGF & 78.72 & 18.48 & 91.82 & -- & 83.48 & \textbf{2.16} & 89.45 & 1.58 & 84.61 & 36.26 \\
+SDM & 78.05 & 22.19 & 91.77 & -- & 83.39 & 2.39 & 90.16 & 1.27 & 84.40 & 26.78 \\
\textbf{SpectralMamba-UNet} & \textbf{81.10} & \textbf{15.31} & \textbf{92.89} & -- & \textbf{83.61} & 2.26 & \textbf{90.99} & \textbf{1.22} & \textbf{86.46} & 23.75 \\
\bottomrule
\end{tabular}
}
\end{table}

\textbf{Quantitative Analysis.} Table~\ref{tab:ablation} evaluates the contributions of spectral decomposition, spatial Mamba modeling, and spectral channel reweighting.
(1) \textbf{Effect of Spectral Decomposition (+Freq).}
Spectral-domain feature extraction improves boundary-sensitive metrics. 
On IA, HD95 decreases from 34.28 to 22.76, with additional gains on Synapse and DRIVE, indicating benefits for both volumetric and tubular structures.
(2) \textbf{Effect of Spatial Mamba.}
Spatial-domain Mamba enhances long-range dependency modeling. 
On DRIVE, DSC increases from 81.96 to 83.07, reflecting improved structural continuity, though gains are not uniform across datasets.
(3) \textbf{Effect of SCR and SGF.}
Adding spectral channel reweighting and guided fusion (+Freq+SCR+SGF) further improves several benchmarks (\textit{e.g.}, DRIVE HD95 reduces to 2.16), highlighting the role of frequency-conditioned decoding.
(4) \textbf{Effect of SDM.}
The +SDM variant combines spectral decomposition with band-wise Mamba modeling, providing additional contextual enhancement. 
The complete SpectralMamba-UNet achieves the best overall performance, demonstrating complementary contributions from spectral modeling and state space dependency learning.

\begin{figure*}[t]
    \centering
    \includegraphics[width=0.8\textwidth]{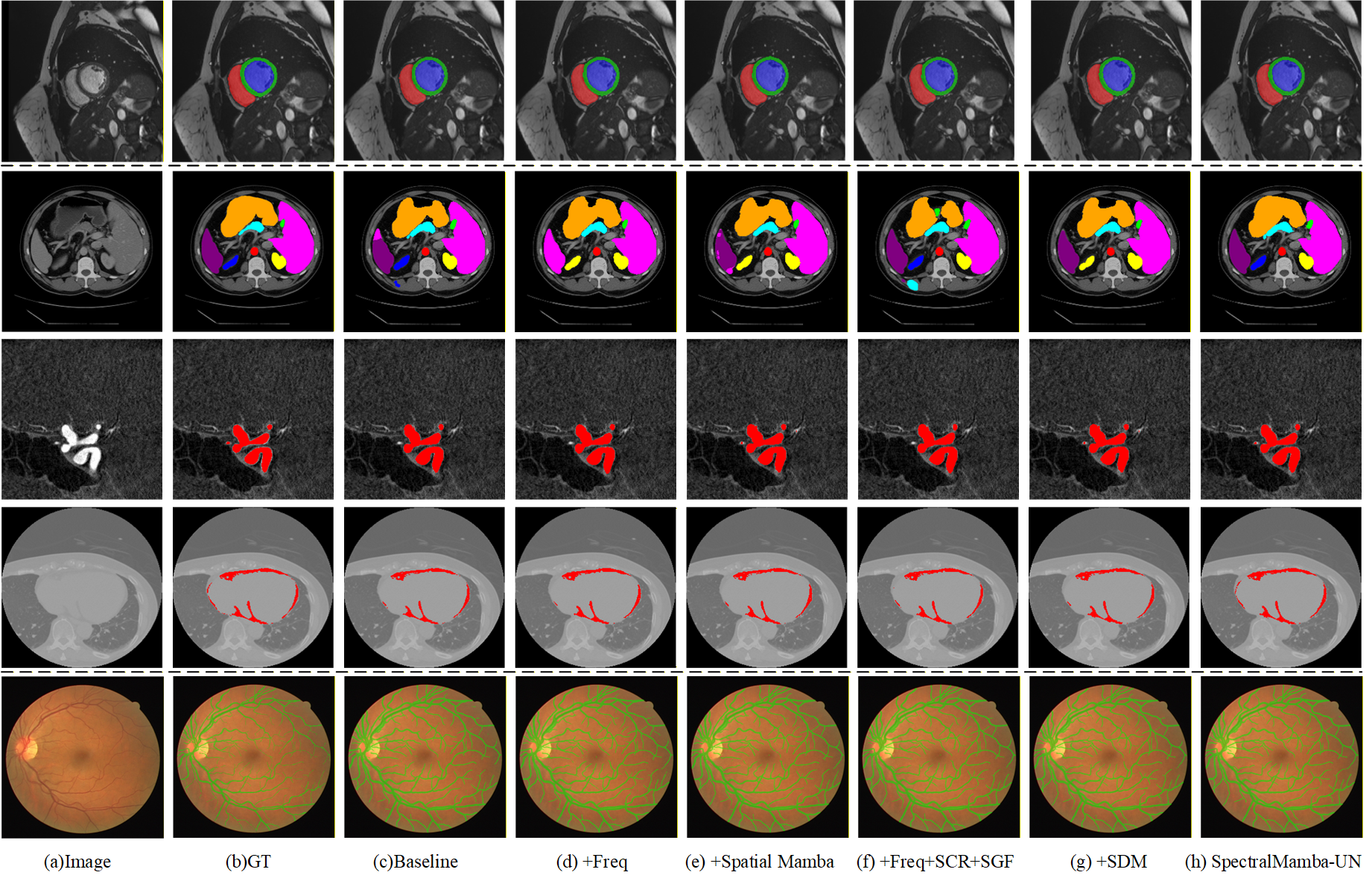} 
    \caption{Qualitative comparison of ablation variants. From left to right: input image, ground truth, Baseline, +Freq, +Spatial Mamba, +Freq+SCR+SGF, +SDM, and the complete SpectralMamba-UNet. 
The full model produces clearer boundaries and improved structural continuity across datasets.}
\vspace{-10pt}
    \label{fig:AB_visual}
\end{figure*}

\textbf{Qualitative Analysis.}
Fig.~\ref{fig:AB_visual} shows progressive refinement. 
The baseline produces coarse boundaries and fragmented structures. 
Spectral decomposition sharpens thin regions, while spatial Mamba improves global coherence. 
The full model integrates these advantages, yielding clearer boundaries and more consistent topology.

\section{Conclusion}

We presented SpectralMamba-UNet, a frequency-disentangled state space framework for medical image segmentation. 
The proposed method decomposes features into complementary spectral components via SDM, adaptively reweights frequency responses through SCR, and performs frequency-guided multi-scale fusion using SGF. 
By coordinating spectral disentanglement with state space modeling, the framework enhances structural consistency and boundary precision. 
Experiments on five public benchmarks show consistent improvements over strong CNN-, Transformer-, and Mamba-based baselines. 
These findings suggest that integrating frequency-domain analysis with state space modeling is a promising and generalizable direction for medical image segmentation.

\section*{Acknowledgments}
This work was supported by the National Natural
Science Foundation of China (62006165) and the Natural Science Foundation of Sichuan Province (2025ZNSFSC1477).

\bibliographystyle{unsrt}  


\end{document}